\documentclass{article} 
\usepackage{nips13submit_e,times}
\usepackage{hyperref}
\usepackage{url}
\usepackage{graphicx}
\usepackage{verbatim}

\title{Learning Paired-associate Images with An Unsupervised Deep Learning Architecture}

\author{
Ti Wang and Daniel L. Silver \\
Jodrey School of Computer Science\\
Acadia University\\
Wolfville, NS, Canada B4P 2R6 \\
\texttt{danny.silver@acadiau.ca} \\
}

\nipsfinalcopy 

\begin{document}

\maketitle

\begin{abstract}
This paper presents an unsupervised multi-modal learning system that learns associative representation from two input modalities, or channels, such that input on one channel will correctly generate the associated response at the other and \emph{vice versa}. 
In this way, the system develops a kind of supervised classification model meant to simulate aspects of human associative memory. 
The system uses a deep learning architecture (DLA) composed of two input/output channels formed from stacked Restricted Boltzmann Machines (RBM) and an associative memory network that combines the two channels. 
The DLA is trained on pairs of MNIST handwritten digit images to develop hierarchical features and associative representations that are able to reconstruct one image given its paired-associate.
Experiments show that the multi-modal learning system generates models that are as accurate as back-propagation networks but with the advantage of a bi-directional network and unsupervised learning from either paired or non-paired training examples.

\end{abstract}

\section{Introduction}
Humans learn knowledge from the environment by data that is provided in several forms, or \emph{modalities}, such as audio and visual signals.
Psychologists define multi-modal learning as learning new knowledge from multiple sensory modalities~\cite{paivio:1990mental}.
Researchers have shown that people's understanding of new concepts is enhanced with mixed-modality knowledge representations~\cite{mayer:2009multimedia}.
The human brain has adapted to fuse associated sensory signals so as to learn more effectively and efficiently.
The long-term goal of this research is to develop a learning system that simulates aspects of the multi-modal learning ability of humans. 
In particular, we investigate unsupervised learning methods that can create a model capable of generalization and classification from one input or output modality to another (eg. from visual to verbal). 
We are interested in how this can be done without resorting to any form of supervised learning that suffers from the need for labeled examples. 

Deep learning is a sub-area of machine learning, which typically uses Restricted Boltzmann Machines (RBM), a type of stochastic associative artificial neural network (ANN), to develop a multi-layer generative models~\cite{Hinton:2007layers}.
Deep learning architectures, or DLA, provide an exciting new substrate upon which to explore new computational and representational models of how knowledge can be acquired, consolidated and used~\cite{Bengio:2009}.  Prior work has investigated the use of DLAs and unsupervised learning methods to develop models for a variety of purposes including auto-associative memory, pattern completion, and clustering as well as generalization and classification~\cite{Hinton:2006}.

This paper takes a first step toward developing a multi-modal learning system by examining a DLA that is capable of learning paired-associate images at two input modalities (channels).
The DLA must reconstruct the matching image at channel A when it observes it's paired image at channel B, and \emph{vice versa}.
By doing so the system uses unsupervised learning to develop an associative memory model that performs a form of classification from one channel to another.
Additionally, this DLA can learn not only paired-associate examples, but also non-paired independent examples at each sensory modality. 
Experimentation shows quantitatively and qualitatively that the system generates models that accurately generates associated images as compared to models developed using traditional supervised back-propagation networks.

\section{Background}
Artificial neural networks (ANN) are widely used to solve classification problems such as image and speech recognition, however many do not work in the same fashion as the human nervous system. 
For example, back-propagation ANNs are good for modeling complex mapping relations between input and output data, but are not as good for reconstructing, or recalling a pattern.
Humans have the ability to recover complete information from partial information; this is referred to as associative memory~\cite{Gerrig:2007}. 
When a child watches a tennis game, he or she learns the appearance of the tennis ball and the racket. Next time when the child sees a picture of a tennis ball, the child may recall an image of a racket and of the game.  Associations are clearly a major part of  learning about the world. 

Associative ANNs are inspired by cognitive psychology and are designed to mimic the way that collections of biological neurons may store and recall associative memories~\cite{Palm:2013}.
Geoffrey Hinton, University of Toronto, advocates using Boltzmann Machine associative networks to simulating human brain structure. 
After a Boltzmann Machine has been trained on a set of patterns, it has the ability to reconstruct any one of those  patterns from a partial or noisy pattern.
However, learning is slow in large Boltzmann Machines because of the many weights in a fully connected network and the iterative sampling of node activities required for each weight update. 

\subsection{Restricted Boltzmann Machine}

\begin{figure}[t]
\begin{minipage}[t]{0.4\linewidth}
\centering
\includegraphics[width=2.0in,height=1.3in]{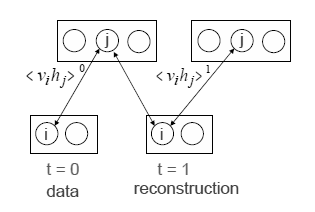}
\caption{RBM Training Process} 
\label{RBMtrain}
\end{minipage}%
\begin{minipage}[t]{0.6\linewidth}
\centering
\includegraphics[width=3.2in,height=1.5in]{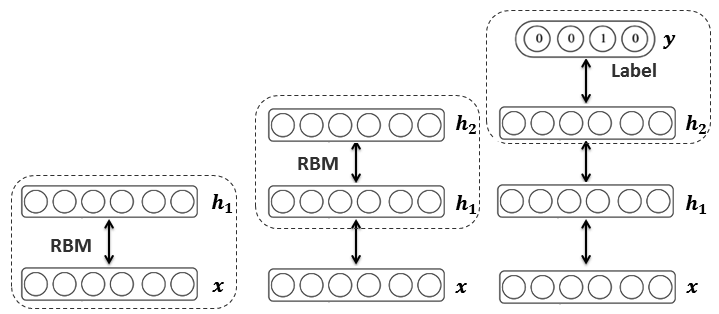}
\caption{Stacking Multi-level RBMs} 
\label{rbm_stack}
\end{minipage}
\end{figure}

A Restricted Boltzmann Machine (RBM) is a variant of a BM that is meant to overcome long training times by limiting the number of connections in its network and using a modified learning algorithm. 
RBMs have both visible and hidden layers of neurons just like BMs, however there are no intra-layer connections, so they can be characterized as a bipartite graph (see Figure~\ref{RBMtrain})~\cite{Hinton:2006}.
When settling to equilibrium, neuron $h_j$ turns on with the probability $p_j= \frac{1}{1 + exp(- b_j  - \sum_i w_{ij}v_i)}$, and neuron $v_i$ turns on with the probability $p_i=\frac{1}{1 + exp(- b_i - \sum_j w_{ij}h_j)}$.
The states $v_i$, $h_j$ of neuron $i$ and $j$ keep changing with probabilities $p_i$ and $p_j$.
The system computes the activation energy $E = - \sum_i b_i v_i - \sum_j b_j h_j - \sum_i \sum_j v_i h_j w_{ij} $ where $b_i$ and $b_j$ are the bias terms for their respective nodes~\cite{Hinton:1986}.  
The global energy $E$ will be reduced more  quickly in an RBM compared to a BM because of the reduced number of connections.
The goal of training is to modify the weights of the network to establish low energy states that correspond with training patterns at the visible nodes.
Similar input patterns will have energy states closer to each other, whereas two orthogonal patterns (e.g. patterns that share few common pixels) will have energy states more distant from each other.

The method of weight update we use for this research is called Contrastive Divergence,  or CD~\cite{Hinton:2006}.
The weights of the network are initialized to small random values.  
When training data $x_i$ is given to the visible neuron $v_i$, the RBM clamps the states of visible neurons and frees the states of hidden binary neuron $h_j$ (see Figure~\ref{RBMtrain}). 
Each weight $w_{ij}$ of the RBM is updated as per the following formula
$\Delta w_{ij} = \eta(< v_i h_j >^0 - < v_i h_j >^1)$, where $\eta$ is the learning rate,  $< v_i h_j >$ is the expectation over all possible pairs of visible and hidden node values, and the 0 and 1 superscripts indicate the expectation based on the training example and its reconstruction, respectively.
This equation approximates the gradient of the log probability of a training example with respect to a weight. 
Weight $w_{ij}$ is updated until the global energy $E$ reduces below a threshold. 
With probability $p_i$, neuron $i$ will then reconstruct the input data $x_i$. 
After training, the hidden layer weights of the RBM will have learned the feature distribution of the input space, that is $w_{ij}$ is equal to the probability of feature $h_j$ given input $v_i$.

To test its ability to recall a pattern, the RBM is presented with all or some of the inputs $x_i$ of a test example at its visible units $v_i$.  These cause activations at each of the hidden units $h_j$ as described above, and then the visible units are freed to  generate new activations.  If training has been successful, the reconstructed outputs at $v_i$ are close to the complete pattern of the original test example. 

\subsection{Deep Learning Architectures}

Humans tend to organize ideas and concepts hierarchically~\cite{Gouws:2012}.
Abstract concepts are learned and recalled through the composition of simpler concepts~\cite{Bengio:2009}.
This approach makes sense in a world where most objects are made from parts which are in turn composed of smaller features.
For instance, a car is a combination of smaller parts like wheels and a frame. 
And a wheel is made up of smaller features like a tire and a rim. 
Neuroscience studies have confirmed that this compositional structure can be seen in the human nervous system. The mammalian brain uses a deep learning architecture with multiple levels of abstraction corresponding to different areas of the neocortex \cite{serre:2007}. 

\emph{Deep learning architectures},  or DLA, is a sub-area of machine learning that places heavy emphasis on hierarchical composition and unsupervised learning methods.  
DLAs can be developed by stacking layers of RBMs one on top of another~\cite{Hinton:2006}.
They have been successfully used to develop models for recognizing hand-writing images of digits in a manner that  simulates the human visual cortex~\cite{Hinton:2007layers}
RBM-based DLA systems are capable of doing unsupervised \emph{clustering} of unlabeled data based on a hierarchy of features. 
As shown in Figure~\ref{rbm_stack}, the hidden layer of one RBM can be used as the input layer for a higher level RBM~\cite{Bengio:2009}.
The highest level features can be used to achieve classification, if so desired.
Subsequently, researchers feel that DLAs develop a hierarchy of features in a fashion similar to the mammalian brain.

DLAs present a new way at looking at systems that learn.  
Deep architectures can be used as an auto-encoder to model high-dimensional data, such as images and audio~\cite{Hinton:2010audio}.
Bengio reports that deep architectures are more expressive than shallow ones by analyzing the depth-breadth trade-off of architecture representation~\cite{Bengio:2007}.
Perhaps most importantly, deep learning methods learn representative hierarchies directly from the data~\cite{Bengio:2009}.
This is in contrast to approaches such as convolutional networks that use receptive fields and modified back-propagation methods that rely heavily on known topological characteristics of the input space~\cite{Lecun:2007}. 

\section{Multi-modal Learning Using an Unsupervised DLA}

\begin{figure}[t]
\begin{minipage}[t]{0.5\linewidth}
\centering
\includegraphics[height=1.6in,width=2.2in]{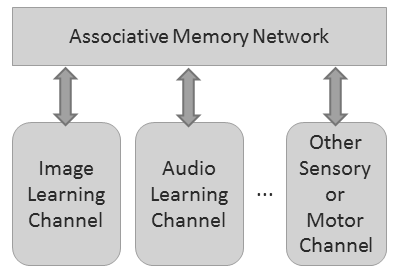}
\caption{Multi-modal data learning system}
\label{multi}
\end{minipage}
\begin{minipage}[t]{0.5\linewidth}
\centering
\includegraphics[height=1.9in,width=2.2in]{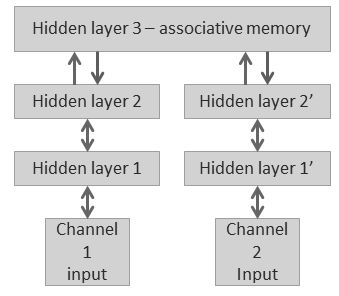}
\caption{Two channels DLA}
\label{DLA}
\end{minipage}
\end{figure}

The objective of this research is to develop a learning system that can memorize and recall multi-channel data using an associative memory network. 
The learning system should be able to recall the pattern from the associative network on one sensory modality given data on another sensory modality. 
The long-term goal of our research is to create a system that can learn concepts using  two or more sensory/motor modalities, such as audio, optical, and vocal (see Figure~\ref{multi}).

\subsection{Learning Paired-Associate Images}

Consider the problem of learning paired-associate images at two input modalities (channels).
We propose to use a DLA network that, after training, will be able to generate a paired image on one channel when prompted with an image on another channel. 
The process is meant to simulate human sensory modalities and associative memory, and to provide insights into how classification can be done using an unsupervised learning approach. 
The learning system is composed of two major parts, a associative memory network and two associative sensory channel networks (see Figure~\ref{DLA}). The sensory channel networks are designed for the recognition and reconstruction of sensory data. The associative memory network ties the sensory channel networks together and simulates the human associative memory.  Both parts can be built using RBMs.

Because of its reduced representation, the recall capacity of an RBM is not as high as a fully-connected BM. 
We have determined that an RBM is unable to recall patterns when only half of the visible neurons are given correct  pattern values~\cite{TiWangThesis2013}.
Thus when an RBM is used as the top associative memory network, additional steps are required after the CD algorithm has completed training.  As per Hinton, the weights of the network require fine tuning~\cite{Hinton:2007layers}.

\begin{figure}[t]
\begin{minipage}[t]{0.7\linewidth}
\centering
\includegraphics[height=1.1in,width=3.8in]{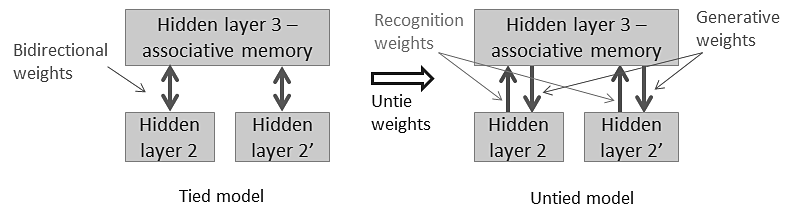}
\caption{Untieing the weights}
\label{untie}
\end{minipage}
\begin{minipage}[t]{0.2\linewidth}
\centering
\includegraphics[height=2.0in,width=1.4in]{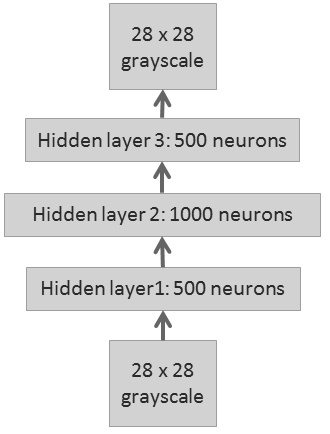}
\caption{BP ANN in Experiment 1}
\label{exp1_bp}
\end{minipage}
\end{figure}

To produce appropriate features at the top layer, the weights of the RBM model need to be fine-tuned. 
However, fine-tuning the bi-directional weights of the RBM may destroy their ability to generate lower level features.
To protect the accuracy of the generative model, it is necessary to untie the weights between the top layer of each channel and the associative memory network layer and create two sets of weights - recognition weights and generative weights (see Figure~\ref{untie})~\cite{Hinton:1995wake,Hinton:2006}.
The recognition weights are used in the bottom-up pass which receives an input pattern and the generative weights are used in the top-down pass to reconstruct an output pattern.  The generative weights are left as trained by the RBM.  The recognition weights are fine-tuned using a back-fitting algorithm, such that the associative memory network can generate a relatively accurate full set of associative memory features with only input from one channel.

To fine-tune channel 1, the recognition weights $w_{ij}$, where $i$ is a neuron in hidden layer 2 and $j$ is a neuron in hidden layer 3, are used as the initial weight values for a gradient descent regression over all paired patterns.
For each training pattern, the posterior probabilities $\{p_i\}$ of hidden layer 2 are used as the input attribute, and the posterior probabilities $\{p_j\}$ of hidden layer 3 are used as the target output.
A new set of posterior probabilities $\{p_j'\}$ for  hidden layer 3 are computed using  $p_j'=\frac{1}{1+exp(-\sum_{i}w_{ij}p_i)}$, and the weights are updated using gradient descent to minimize the error between $\{p_j\}$ and $\{p_j'\}$.
In this way the recognition weights which pass the input signal from sensory channel 1 to the associative memory network are fine-tuned to generate a full set of associative memory features which channel 2 can use to generate the appropriate output.

With back-fitting, the multi-modal DLA should be able to achieve the learning goal that was previously done with supervised learning by Srivastava~\cite{Nitish:2012}.
Without supervised learning between the two channels, the performance of the DLA is unlikely to exceed that of a traditional BP ANN approach; however, we do expect it to do as well.
The hierarchical feature learning of the sensory channels and the back-fitting of the recognition weights are expected to make up for the shortcomings of purely unsupervised learning approach that we are taking.

\subsection{Impact of Learning Non-paired Patterns}
\label{nonp}
Sensory data does not always come in pairs in real life.
For example, one can see a cat meowing, see an image of a cat, or hear meowing without seeing a cat.
In this case, the sound ``meow" is the audio signal and the image of the cat is the visual signal.
These two sensory channels can come together to allow paired-associate learning, but their individual channel representations can be learned and improved upon separately.
We propose that learning each sensory modality with non-paired examples will help to improve the associative memories ability to generate the correct image on one channel when given its paired-associate on the other.
It would be informative to have an experiment to test the impact on the multi-channel learning system by separately training the sensory channels with non-paired examples.

\section{Empirical Studies}
Three empirical studies were carried out using two different data sets.
The first and third experiments used paired images from the MNIST dataset of handwritten numeric digits.
The second experiment used paired images from a synthetic dataset of numeric digits.
In all experiments, five pairs of odd and even digits were associated with each: 1-2, 3-4, 5-6, 7-8, 9-0.

\subsection{Experiment 1}
{\bf Objective:}
The objective of this experiment is to compare the unsupervised DLA with a supervised BP ANN approach to learning paired-associate images. 
Each learning system is trained such that when a handwritten digit image is provided, the system will generate its paired digit image.

{\bf Material and Methods:}
This experiment uses a dataset of paired MNIST handwritten digits as the learning domain.
The experiment is repeated four times with different training sets, validation sets and test sets.
Each of these datasets contains 1,000 paired-associate examples that are randomly selected from the MNIST dataset.

A deep learning architecture of RBMs is used to develop an unsupervised learning model for the problem. 
The architecture is in accord with Figure~\ref{DLA}. 
Each channel network is composed of two RBM layers, each of which contains 500 hidden neurons.
Hidden layers 1 and 1' and then layers 2 and 2' will develop more abstract features of the original images~\cite{Hinton:2006}.
The associative top layer contains 1,000 neurons.
The unsupervised DLA uses back-fitting to fine-tune the weights of the associative top layer after the CD algorithm training is finished.

When training the DLAs, the training process of each sensory channel stops when the maximum iteration of 60 is reached, and the associative memory network is trained to 100 iterations.
Validation sets are used to monitor the back-fitting to avoid over-fitting.
The odd digit part of a test example is used to test the reconstruction of its corresponding even digit image, and \emph{vice versa}. 

We developed two BP networks to learn the same paired-associate mapping.
One network is trained to map odd digit images to even digits, the other \emph{vice versa}. 
Both BP networks use the architecture shown in Figure~\ref{exp1_bp}. 
The BP networks use the same training set, validation set and testing set as the DLA.
The validation set is used to prevent the BP algorithm from over-fitting to the training set.

The accuracy of reconstruction is measured by testing the output images using Hinton's DLA handwritten digits classification software.
This software is known to classify MNIST dataset of handwritten digits with only 1.15\% errors~\cite{Hinton:2006}.
One can pass the input images and the reconstructed images through \textit{Hinton's classifier} to determine their digit category. The accuracy of the models is then based on the number of correctly paired images. 

\begin{table}[t]\small
\centering  
\begin{tabular}{|c|c|c|c|c|c|c|c|c|c|c|c|}
\hline
&1$\rightarrow$2& 2$\rightarrow$1& 3$\rightarrow$4& 4$\rightarrow$3& 5$\rightarrow$6& 6$\rightarrow$5& 7$\rightarrow$8& 8$\rightarrow$7& 9$\rightarrow$0& 0$\rightarrow$9 &Avg  \\ \hline
DLA &95.25 &95.88 &82.63 &94.63 &92.38 &88.75 &90.5 &79.75 &91.63 &93 &90.74
\\ \hline
BP ANNs &98.0 &72.5 &83.75 &95.13 &90.38 &82.88 &91.13 &82.88 &89.0 &92.5 &88.82
\\ \hline
\end{tabular}
\caption{Accuracy of test set reconstruction (\%)}
\label{exp5_1000}
\end{table}

\begin{figure}[t]
\centering
\includegraphics[height=1.2in]{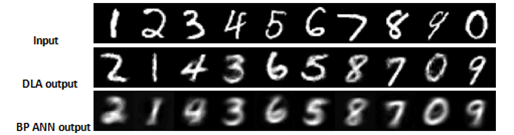}
\caption{Examples of reconstruction results with the DLA and BP ANNs}
\label{exp5_1}
\end{figure}

{\bf Results and Discussion:}
Using Hinton's software, the reconstruction accuracy was checked on the testing set. 
The average results of four replications of the experiments are shown in Table~\ref{exp5_1000}. 
On average, the unsupervised DLA (model 1) generated images that were 90.74\% accurate, and the BP ANNs (model 2) generated images that were 88.82\% accurate.
One can see that the two models did equally well.
This suggests that the unsupervised DLA models are able to achieve the same level of accuracy as the supervised BP approach.

Figure~\ref{exp5_1} shows examples of reconstructed images produced by the DLAs and the BP ANNs. 
One can see that the images generated by the DLAs are clearer than those generated by the BP ANNs. 
We suspect this because the DLA models are able to better differentiate features from noise. 
This will be investigated further in the next experiment.

\subsection{Experiment 2}
{\bf Objective:}
The objective of this experiment is to develop auto-associative models that can overcome noise injected into synthetic training examples. 
An unsupervised DLA with back-fitting and supervised BP ANNs will be developed from a noisy dataset, and the quality of their regenerated images will be compared.

\begin{figure}[h]
\centering
\includegraphics[width=3.2in]{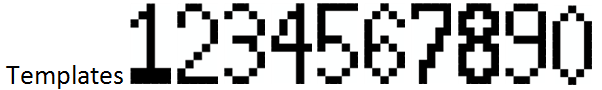}
\caption{Templates of the synthetic dataset}
\label{exp2tem}
\end{figure}

{\bf Material and Methods:}
This experiment uses a synthetic dataset that contains five different sets of 10 x 5 paired images from Figure~\ref{exp2tem}.
10\% random noise was added to each template image to produce 60 instances of each category, or 300 in total.  The first 100 of these images are used as a training set, the next 100 are used as a validation set, while the remaining 100 are used as a test set.

A DLA architecture, in accord with the previous experiment, is used to develop an unsupervised learning model.
Each of the sensory channel layers contains 50 hidden neurons, and the associative top layer contains 100 neurons.
The training process of the sensory channel networks stops when the maximum iteration of 60 is reached; the associative memory network trains for 100 iterations.

As in Experiment 1, two BP networks were developed to learn the same paired-associate mapping.
Both BP networks used an architecture similar to that shown in Figure~\ref{exp1_bp} with 50 neurons in layers 1 and 3 and 100 neurons in layer 2.
The BP networks uses the same training set, validation set and test set as the DLA.

The accuracy of reconstruction was measured by comparing the similarity between the generated images and their corresponding template images for a set of test examples.
The RMSE between the pixels of each reconstructed image and its corresponding template (without noise) was computed to give an average error over all examples (image pixels are normalized to the range [0,1]).

\begin{table}[t]\small
\centering  
\begin{tabular}{|c|c|c|c|c|c|c|c|c|c|c|c|}
\hline
&1$\rightarrow$2& 2$\rightarrow$1& 3$\rightarrow$4& 4$\rightarrow$3& 5$\rightarrow$6& 6$\rightarrow$5& 7$\rightarrow$8& 8$\rightarrow$7& 9$\rightarrow$0& 0$\rightarrow$9 &Avg  \\ \hline
DLA &0.012 &0.071 &0.046 &0.029 &0.004 &0.01 &0.008 &0.0 &0.04 &0.015 &0.032
\\ \hline
BP ANNs&0.162 &0.216 &0.209 &0.081 &0.06 &0.115 &0.135 &0.165 &0.11 &0.106 &0.144
\\ \hline
\end{tabular}
\caption{RMSE of test set reconstruction (out of 1)}
\label{exp5_syn}
\end{table}

\begin{figure}[t]
\centering
\includegraphics[width=3.6in]{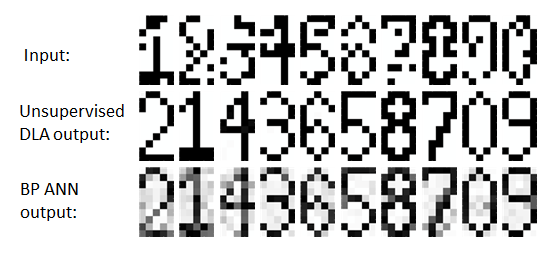}
\caption{Examples of reconstruction results with DLA and BP ANNs}
\label{exp5_3}
\end{figure}

{\bf Results and Discussion:}
The RMSE of the reconstructed images for the test set is shown in Table~\ref{exp5_syn}.
The DLA with back-fitting out-performs the BP networks in generating the images in the presence of noise.
Figure~\ref{exp5_3} shows examples of reconstructed images from the DLA and the BP ANNs.
The generated images from the DLA are quite similar to the template images of Figure~\ref{exp2tem}, while there is significant noise on the generated images from the BP network.
DLAs attempt to probabilistically differentiate features from noises, whereas BP ANNs attempt to map input pixels to output pixels.
Features are formed in BP networks, but they are for the purpose of mapping and not reconstruction of the original images.
Hence a DLA is a better choice if the objective is to construct a noiseless category example as a form of classification.

\subsection{Experiment 3}
{\bf Objective:}
The preceeding experiments used paired-associate examples to develop neural network models, however, sensory data does not always come in pairs in real life.
The objective of this experiment, in accord with Section~\ref{nonp}, is to develop an associative learning system with both paired associative examples and independent non-paired examples.
The experiment is designed to test if the performance of an associative learning system can be improved by separately training the sensory channels with non-paired examples.

{\bf Material and Methods:}

\begin{figure}[t]
\begin{minipage}[t]{0.37\linewidth}
\centering
\includegraphics[height=1.8in]{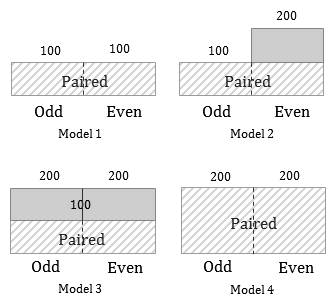}
\caption{The number of examples used to train four models}
\label{4subsets}
\end{minipage}
\begin{minipage}[t]{0.6\linewidth}
\centering
\includegraphics[height=2.2in]{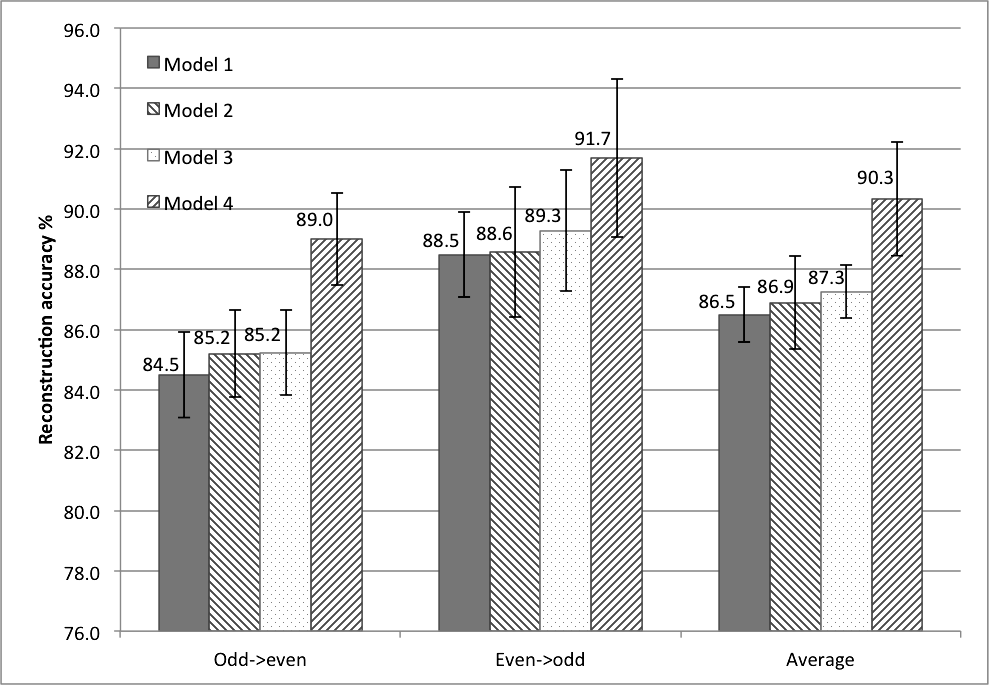}
\caption{Accuracy comparison between four models}
\label{3subsetsbar}
\end{minipage}
\end{figure}

This experiment uses the database of MNIST examples as in Experiment 1.
The experiment is repeated four times with different training sets, validation sets and test sets.
For each repetition, four models are built using the same architecture but with different amounts of training examples.
The first model is built with 100 paired-associate examples.
The second model is built with 100 paired-associate examples, and 100 non-paired examples of even digit images.
The third model is built with 100 paired-associate examples, 100 non-paired examples of even digit images, and 100 non-paired examples of odd digit images.
The last model is built with 200 paired-associate examples.
Figure~\ref{4subsets} shows the number of paired and non-paired examples in each training set.
All the odd digits images are used to train the odd channel and all the even digits are used to train the even channel, but only the paired-associate examples are used to develop the associative memory.

The four models use the same 3-layered architecture, parameters, validation sets and test sets as in Experiment 1.
While doing back-fitting, validation sets are used to monitor overfitting. 
Test sets are used to examine the associative learning performance of the learning system.
The odd digits are used to test the recall of even digits, and \emph{vice versa}.
The recalled images are classified by \textit{Hinton's classifier} to examine the accuracy of the models.

{\bf Results and Discussion:}

The performance (averaged over four repetitions) of the four models at recalling even digits from odd digits, odd digits from even digits, and the average of them are shown in Figure~\ref{3subsetsbar}; the error bars represent the 95\% confidence over the repeated studies.
The mean accuracy increases marginally (the error bars show that the improvements are not significant) from model 1 to model 3, which means that using non-paired examples to better develop one of the channels representation may  improve the overall performance of an associative learning system.
We conjecture that this is because both the recognition weights and the generative weights of this channel are optimized.
Improving the recognition weight performance of the odd digits channel will provide better features to the associative memory network to generate the corresponding even digits.
Better generative weights for the odd digits channel will generate more accurate odd digits when even digits are provided.
In general, this result suggests that improving one of the sensory channel networks of a multi-channel learning system which contains more than two channels will improve any recall that involves that channel.

It is also important to note that the reconstruction accuracy clearly increases from model 3 to model 4. This demonstrates that using more paired-associate examples to develop the associative memory network can improve the performance of the system over the equivalent number of non-paired examples.   
In a system with three or more channels we conjecture that paired-associate examples for any two channels will be of benefit to the entire associative memory network. 

\section{Conclusion}
This paper presents recent work on an unsupervised multi-modal learning system that can develop an associative memory structure that combines two input/output channels.
Our long-term goal is to develop learning systems that are able to learn conceptual representations from multiple sensory input and/or motor output modalities in a manner similar to humans. 

We have demonstrated an unsupervised deep learning architecture (DLA) that can reconstruct an image of a MNIST handwritten digit from another paired handwritten digit.  
The system develops a kind of supervised classification model meant to simulate aspects of human associative memory. 
The DLA is formed with stacked Restricted Boltzmann Machines (RBM) and trained with the Contrastive Divergence (CD) algorithm.
The RBM associative memory network that ties the input/output channels together requires refinement using a back-fitting technique to increase the recall accuracy when only 50\% of its visible neurons are available from one channel.
Experimentation shows quantitatively (using an independent classification method) and qualitatively (by viewing the generated images) that the system develops models that are able to reconstruct accurate paired images as compared  to supervised back-propagation network models and have the advantage of unsupervised learning from either paired or non-paired training examples.  

In future work, different types of sensory data will be used to train the multi-modal learning system, such as audio signals.
Furthermore, we are interested in knowledge transfer in DLAs using unsupervised methods for learning new tasks and new modalities.

\small
\bibliographystyle{plain}
\bibliography{thesis}

\begin{thebibliography}{10}

\bibitem{Bengio:2009}
Yoshua Bengio.
\newblock Learning deep architectures for ai.
\newblock {\em Found. Trends Mach. Learn.}, 2(1):1--127, January 2009.

\bibitem{Bengio:2007}
Yoshua Bengio and Yann Lecun.
\newblock {\em {Scaling learning algorithms towards AI}}.
\newblock MIT Press, 2007.

\bibitem{Hinton:2010audio}
Li~Deng, Michael~L. Seltzer, Dong Yu, Alex Acero, Abdel rahman Mohamed, and
  Geoffrey~E. Hinton.
\newblock Binary coding of speech spectrograms using a deep auto-encoder.
\newblock In Takao Kobayashi, Keikichi Hirose, and Satoshi Nakamura, editors,
  {\em Interspeech}, pages 1692--1695. ISCA, 2010.

\bibitem{Gerrig:2007}
Richard~J. Gerrig and Philip~G. Zimbardo.
\newblock {\em Psychology and Life}.
\newblock MyPsychLab Series. Pearson/Allen and Bacon, 2007.

\bibitem{Gouws:2012}
Stephan Gouws.
\newblock Deep unsupervised feature learning for natural language processing.
\newblock In {\em Proceedings of the 2012 Conference of the North American
  Chapter of the Association for Computational Linguistics: Human Language
  Technologies: Student Research Workshop}, NAACL HLT '12, pages 48--53,
  Stroudsburg, PA, USA, 2012. Association for Computational Linguistics.

\bibitem{Hinton:2007layers}
Geoffrey~E. Hinton.
\newblock {Learning multiple layers of representation}.
\newblock {\em Trends in Cognitive Sciences}, 11:428--434, 2007.

\bibitem{Hinton:1995wake}
Geoffrey~E. Hinton, Peter Dayan, Brendan~J. Frey, and Radford~M. Neal.
\newblock The wake-sleep algorithm for unsupervised neural networks.
\newblock {\em Science}, 268(5214):1158--1161, 1995.

\bibitem{Hinton:2006}
Geoffrey~E. Hinton and Simon Osindero.
\newblock A fast learning algorithm for deep belief nets.
\newblock {\em Neural Computation}, 18:2006, 2006.

\bibitem{Hinton:1986}
Geoffrey~E. Hinton and Terrence~J. Sejnowski.
\newblock Parallel distributed processing: explorations in the microstructure
  of cognition, vol. 1.
\newblock chapter Learning and relearning in Boltzmann machines, pages
  282--317. MIT Press, Cambridge, MA, USA, 1986.

\bibitem{mayer:2009multimedia}
Richard.E. Mayer.
\newblock {\em Multimedia Learning}.
\newblock Cambridge University Press, 2009.

\bibitem{paivio:1990mental}
Allan. Paivio.
\newblock {\em Mental representations}.
\newblock Oxford University Press, Incorporated, 1990.

\bibitem{Palm:2013}
G.~Nther Palm.
\newblock Neural associative memories and sparse coding.
\newblock {\em Neural Netw.}, 37:165--171, January 2013.

\bibitem{Lecun:2007}
Marc'Aurelio Ranzato, Y~lan Boureau, and Yann Lecun.
\newblock Sparse feature learning for deep belief networks.
\newblock In {\em NIPS-2007}, 2007.

\bibitem{serre:2007}
Thomas Serre, Gabriel Kreiman, Minjoon Kouh, Charles Cadieu, Ulf Knoblich, and
  Tomaso Poggio.
\newblock A quantitative theory of immediate visual recognition.
\newblock {\em Prog Brain Res}, pages 33--56, 2007.

\bibitem{Nitish:2012}
Nitish Srivastava and Ruslan Salakhutdinov.
\newblock Multimodal learning with deep boltzmann machines.
\newblock In P.~Bartlett, F.C.N. Pereira, C.J.C. Burges, L.~Bottou, and K.Q.
  Weinberger, editors, {\em Advances in Neural Information Processing Systems
  25}, pages 2231--2239. 2012.

\bibitem{TiWangThesis2013}
Ti~Wang.
\newblock {\em Classification Via Reconstruction Using A Multi-Channel Deep
  Learning Architecture}.
\newblock Masters Thesis, Acadia University, Wolfvillle, NS, Canada, 2013.

\end{thebibliography}
\end{document}